\documentclass[10pt]{article}
\usepackage{graphicx}
\usepackage{fullpage}

\usepackage{amsthm}
\usepackage{subcaption}
\usepackage{latexsym}
\usepackage{helvet}
\usepackage{courier}
\usepackage{amsmath}
\usepackage{amssymb}
\usepackage{multirow}
\usepackage{mdwlist}
\usepackage{setspace}
\usepackage{tabularx}
\usepackage[noend]{algorithmic}
\usepackage{algorithm}
\usepackage{multirow}
\usepackage{bm}

\usepackage{soul}
\usepackage{graphicx}
\usepackage{graphics}
\usepackage{times}
\usepackage{latexsym}
\usepackage{helvet}
\usepackage{courier}
\usepackage{amsmath}
\usepackage{amssymb}
\usepackage{url}
\usepackage[noend]{algorithmic}
\usepackage{algorithm}
\usepackage{multirow}
\usepackage{mathtools}
\usepackage{enumitem}
\usepackage{color}
\usepackage{microtype}
\usepackage{tabularx}

\usepackage{amsthm}

\title{Towards a Decentralized, Autonomous Multiagent Framework for Mitigating Crop Loss}
\author{
  Roi Ceren\\
  Dept. Computer Science\\
  University of Georgia\\
  \texttt{roi.ceren@gmail.com}
  \and
  Shannon Quinn\\
  Dept. Computer Science\\University of Georgia\\
  \texttt{spq@uga.edu}
  \and
  Glen Rains\\
  Dept. Entomology\\
  University of Georgia\\
  \texttt{grains@uga.edu}
}
\date{October 2018}

\begin{document}

\maketitle

\begin{abstract}
    We propose a generalized decision-theoretic system for a heterogeneous team of autonomous agents who are tasked with online identification of phenotypically expressed stress in crop fields. This system employs four distinct types of agents, specific to four available sensor modalities: satellites (Layer 3), uninhabited aerial vehicles (L2), uninhabited ground vehicles (L1), and static ground-level sensors (L0). Layers 3, 2, and 1 are tasked with performing image processing at the available resolution of the sensor modality and, along with data generated by layer 0 sensors, identify erroneous differences that arise over time. Our goal is to limit the use of the more computationally and temporally expensive subsequent layers. Therefore, from layer 3 to 1, each layer only investigates areas that previous layers have identified as potentially afflicted by stress. We introduce a reinforcement learning technique based on Perkins' Monte Carlo Exploring Starts for a generalized Markovian model for each layer's decision problem, and label the system the Agricultural Distributed Decision Framework (ADDF). As our domain is real-world and online, we illustrate implementations of the two major components of our system: a clustering-based image processing methodology and a two-layer POMDP implementation.
\end{abstract}

\section{Introduction}
\label{sec:intro}

The low-cost availability of imaging technology has given rise to the rapidly developing field of precision agriculture, often marked by the use of multispectral image collection via autonomous uninhabited aerial vehicles (AUAVs)~\cite{precag}. As an example, recent efforts combining AUAVs, normalized difference vegetation index (NDVI) imaging, and environmental barometric and water potential sensors have been used to create efficient autonomous systems for targeted crop field watering~\cite{farmbeats}. Additionally, many targeted image processing systems have been developed for the purpose of specific disease identification based on phenotypic expression, such as lesions, browning, and tumors~\cite{plant_detection}.

While precision watering techniques have dramatically improved yields for large-scale farms, the advent of autonomous intervention for disease propagation is nascent~\cite{plant_detection}. While some generalized models exist to detect these stresses, they have not been introduced to the distributed autonomous systems as in precision watering. To that end, we adopt the problem of identifying and predicting the onset of stresses (pest and pathogen) in crop fields via environmental sensor and image data, taken at various resolutions throughout a growing season. We factor the distribution of functional capabilities of our physical system into four distinct layers, comprised of satellites (layer 3), AUAVs (L2), autonomous uninhabited ground vehicles (AUGVS, L1), and static ground-level sensors (L0). Generally, the output of each layer (excluding L0) is used to inform the decision making of the layer below it by raising a \textit{call-to-action}, wherein the layer believes a stress is occurring based on phenotypic expression that differs in a geographical location.

\begin{figure} [htb]
\label{fig:intro_layers}
\centerline{\includegraphics[width=8cm]{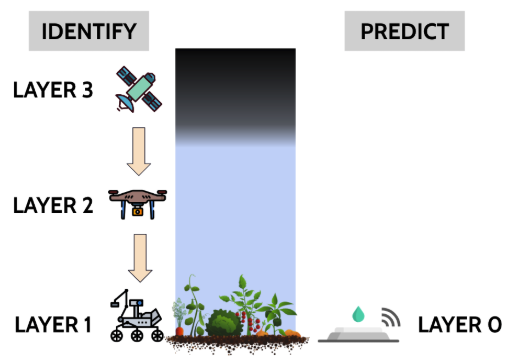}}
\caption{Layer composition of physical sensor modalities in ADDF. Images and Layer 0 data are used in each layer to inform subsequent layers of possible stress, where UGVs are constrained by UAVs, and UAVs are constrained by satellites.}
\end{figure}

A common challenge of real-world problem domains, particularly in the agricultural domain, is the constraint on sample availability for machine learning. Since we are attempting to uncover the true state of stress in a crop field without prior knowledge, we propose a \textit{model-free} exploration of policies \textit{a la} Perkins' Monte Carlo Exploring Starts (MCES) for Partially Observable Markov Decision Processes (POMDPs), labeled MCES-P~\cite{perkins}. MCES-P iterates over memory-less policies that directly map actions to observations, instead of beliefs of the state~\cite{wiering}. As a first best effort towards our goal, we assemble our sensor modalities into a heterogeneous team, utilize an image processing technique to extract potentially stressed sectors, and learn policies that map these observations of phenotypic deviations to calls for intervention.

This work is divided into the following sections. Section~\ref{sec:related} covers the related topics in precision agriculture. Section~\ref{sec:background} introduces the necessary concepts for the algorithms used in Sec.~\ref{sec:method}. Before covering the framework, Sec.~\ref{sec:domain} describes the problem domain, including descriptions of our crop fields and the arrangement of our available physical sensor modalities. We then test prototypical experiments of our real-world domain in Sec.~\ref{sec:results}.

\section{Related Work}
\label{sec:related}

Our approach falls under the body of work characterized by the category of precision agriculture, tackled by a variety of fields, including agriculture, agronomics, computer science, robotics, engineering, and physics. In particular, the relevant subtopics we explore include disease detection, nutrient deficiency, and insufficient water potential. This data provides a basis for precision agro-management, such as through spot spraying, targeted water irrigation and nitrogen application.

The most recent advance in precision agriculture is the FarmBeats initiative driven by Microsoft AI~\cite{farmbeats}, in which a variety of network-accessible sensor modalities, including soil water potential sensors and AUAVs, are arranged to provide automated and targeted water intervention. This methodology is powerful for tackling stresses due to underwatering, but is incapable of detecting the presence of pathogens, pests and nutrient deficiency, which express themselves phenotypically. 

Concerning the goal of disease identification and intervention, the wide array of contemporary efforts leverage phenotypic expressions of stress largely via thermal detection~\cite{khanal} and are often specific to the expression from a specific disease~\cite{plant_detection}. What remains is a generalized model that encompasses the variety of stresses in a model-free way. That is, instead of seeking a particular expression, learn the correlation between erroneous growth patterns, leaf and fruit necrosis or chlorosis, leaf spots, leaf striations and wilting (as caused by stress) and the available image and environmental data. 

\section{Background}
\label{sec:background}

In this section we cover the state of the art on the two major components of our work: modeling the temporal evolution of systems via image processing and employing model-free reinforcement learning in Markov models.

\subsection{Image Processing}
\label{sec:bg_img}

The availability of high-dimensional multispectral image data of crop fields in the last few years~\cite{landsat} has dramatically increased the development of computational systems designed to analyze and interpret crop image data. In parallel, NDVI was established as a powerful metric for image data, as it computes visual attributes of crop fields while eliminating non-vegetative properties~\cite{ndvi}.

The field of crop field temporally-evolving image processing via NDVI imaging is a nascent and quickly growing field~\cite{landsat_methods}. Contemporary work largely focuses on retrospective curve-fitting, as in time series analysis on Advanced Very-High-Resolution Radiometer (AVHRR)~\cite{avhrr} and Moderate Resolution Imaging Spectroradiometer (MODIS)\cite{modis} data, with several focusing on the root-mean-squared deviation (RMSD) metric over pairwise  pixel differences as an image comparison methodology~\cite{landsat}.

As our system is online, and therefore must make immediate estimates of possibly early or ongoing crop stress, retrospective models do not satisfy our needs. Therefore, we instead focus on adapting these methods to online settings, exploring methodologies using pairwise image differences as a sufficient metric.

\subsection{Model-Free Markov Models}
\label{sec:bg_pomdp}

Monte Carlo Exploring Starts for POMDPs (MCES-P)~\cite{perkins} extends MCES for MDPs~\cite{sutton}, an online implementation of reinforcement learning that explores locally neighboring policies. Algorithm~\ref{alg:MCESP} describes the MCES-P process.

\begin{algorithm}[h]
  \caption{MCES-P}
  \label{alg:MCESP}
  \begin{algorithmic}[1]
    \REQUIRE Q-value table initialized and initial policy, $\pi_i$, that is greedy w.r.t Q-values; a maximum number of trajectories, $k$, to explore; $\alpha$, $\epsilon$
    \STATE $c_{o, a} \gets 0$
    \STATE $m \gets 0$
    \REPEAT
    \STATE Choose source observation history, $o$, and $a$
    \STATE Modify $\pi_i$ to $\pi_i \gets (o, a)$
    \STATE Generate trajectory, $\tau$, according to $\pi_i \gets (o, a)$
    \STATE $Q^{\pi_i}_{o, a} \gets (1-\alpha(m, c_{o, a})) \cdot Q^{\pi_i}_{o, a} +\alpha(m, c_{o,a}) \cdot R_{post-o}(\tau)$
    \STATE $c_{o, a} \gets c_{o, a} + 1$
    \IF {$\max_{a'} Q^{\pi_i}_{o, a'} > Q^{\pi_i}_{o, \pi_i(o)} + \epsilon(k, c_{o,a}, c_{o,\pi_i(o)})$}
    \STATE $\pi_i(o) \gets a'$
    \STATE $m \gets m + 1$
    \FORALL {$o_i, o_{\bar{i}}, a_i, a_{\bar{i}}$}
    \STATE $c_{o, a} \gets 0$
    \ENDFOR
    \ENDIF
    \UNTIL{termination}
  \end{algorithmic}
\end{algorithm}

\noindent where $\tau$ is an observation, action, and reward $\{(o_i,a_i,r_i)\}_{i=0}^{|\tau|}$.

MCES-P performs round robin transformations over observation sequences and actions via sample action approximation (SAA)~\cite{saa}. After each transformation, MCES-P compares the current transformed policy against the current best policy and, if it dominates, accepts the transformation as the current best policy.

After each $\tau$ is generated, MCES-P performs a Q-learning update to the current value of the observation sequence-action pair with a depreciating learning rate, $\alpha(c) = 1/(c+1)$~\cite{watkins}. When comparing each policy against the current best policy, MCES-P ensures that both policies have been sampled at least $k$ times. To accomplish this, $\epsilon$ is simply defined as $\epsilon(k,p,q) = +\infty$ if $i<k$ or $j<k$, and $0$ otherwise.

\section{Problem Domain}
\label{sec:domain}

\begin{figure} [htb]
\centerline{\includegraphics[width=8cm]{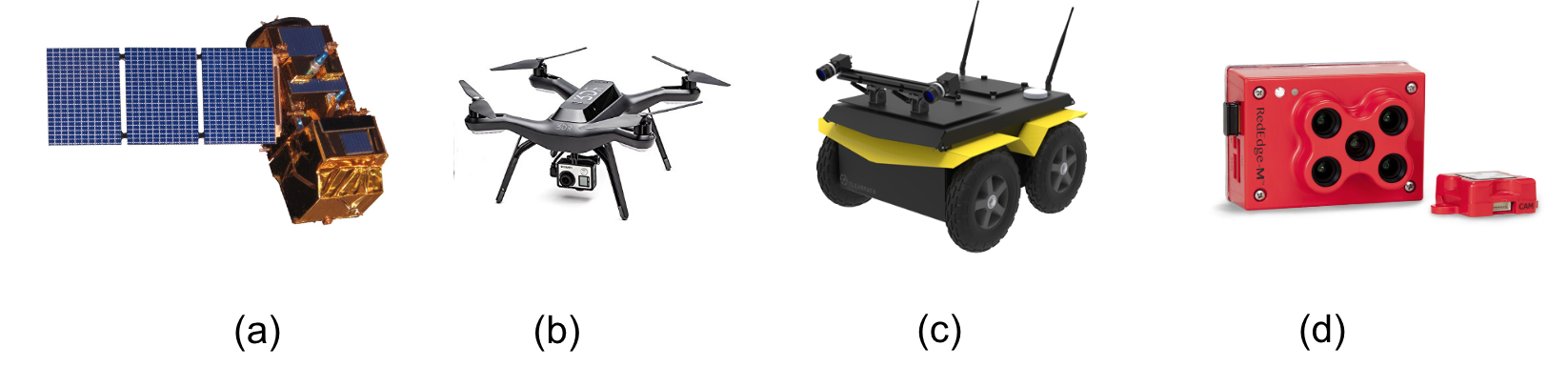}}
\caption{Available sensor modalities: (a) Images from the Sentinal-II, (b) 3DR Solo, (c) ClearPath Jackal, and (d) Micasense Redenge (and others).}
\label{fig:modalities}
\end{figure}

In this work, we propose a novel aggregate framework for utilizing multispectral images and environmental metadata in heterogeneous teams of sensor modalities. In our application, we will a  probability of the presence of disease, nitrogen or water stress, including identifying markers that indicate onset of these stresses. The resultant framework is called the Agricultural Distributed Decision framework (ADDF). ADDF tackles individual and team learning and planning, as well as approaches for accomplishing individual tasks, representing our task of organizing satellites, autonomous uninhabited aerial vehicles (AUAVs), autonomous uninhabited ground vehicles (AUGVs), and ground-level environmental sensors, specified in Fig.~\ref{fig:modalities}.

Beyond sensor modalities, the environmental domain (used in Sec.~\ref{sec:results_image} and simulated in Sec.~\ref{sec:results_team}) is embodied by large-scale peanut crop fields in Tifton, GA, managed by the Department of Agriculture at the University of Georgia, Tifton campus. Extant stresses include crop field erosion, damage done by local fauna, and several introduced stresses including fungus and pathogen introductions resulting in crop blight and lesions.

The functional capabilities of our sensor modalities are defined as follows:

\begin{itemize}
    \item \textbf{L3}: The highest level will be comprised of satellites which collect images of the crops on average every week. The Sentinel 2a and 2b satellites will be used for this layer of information. The data from these satellites are available for free and the ESA also provides a tool box for processing data collected. This level is passive, as it is not directly controlled, instead limited by a fixed temporal component of passing over the same location once every 5 days. Image data in this layer has a multi-spectral resolution of 10 meters per pixel  and a thermal resolution of 20 m per pixel.
    \item \textbf{L2}: Layer 2 represents the multispectral, high dimensional image data taken by AUAVs. While AUAVs have a very high degree of control and speed of execution with limited challenge in pathing, images are taken above their targets. Images generally have a resolution of 1-3 centimeters per pixel depending on the height of the AUAV and the spatial resolution of the camera.
    \item \textbf{L1}: Like L2, this layer represents a controllable physical subsystem, but is executed by AUGVs. Pathing is significantly more challenging and, even given high resolution geolocation sensors, may require exploration for the AUGV to reach its target. Though targeted exploration is more computationally costly and slower, image data can be collected from a variety of perspectives (individual leaves, fruit, flowers, whole plants) and have a similar resolution to L2, albeit from alternative angles.
    \item \textbf{L0}: Although often not directly measuring features due to the onset of stress, we include a layer representing an array of ground sensors which measure air and soil moisture, ground temperature, and relative humidity. This helps acquire information valuable for predictive learning approaches that enhance the data model beyond the other layers’ image data, enhancing model accuracy by capturing features that the resultant stresses are conditionally dependent on.
\end{itemize}

\begin{figure} [htb]
\centerline{\includegraphics[width=8cm]{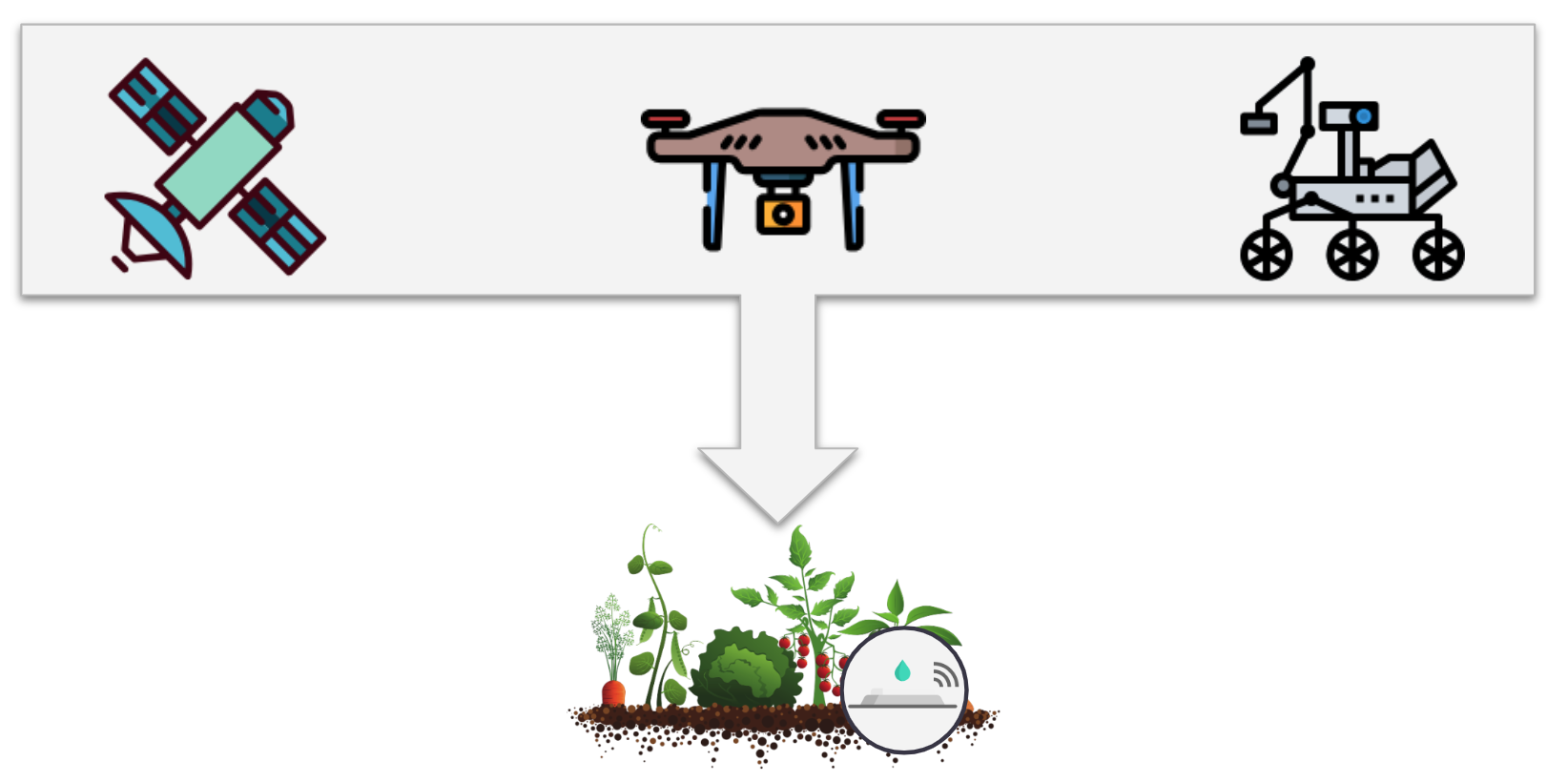}}
\caption{Arrangement of physical sensor modalities. Layers 3, 2, and 1 delegates to subsequent layers, and all layers take image data and collect environmental data from layer 0.}
\label{fig:domain_arrangement}
\end{figure}

As Fig.~\ref{fig:domain_arrangement} illustrates, layers 3, 2, and 1 collect image data at various resolutions and spectra of the crop field, while also collecting environmental data from the layer 0 sensors. Each layer is able to instantly communicate with a centralized server via a wireless network infrastructure. This centralized server then performs analysis and reasoning on the collected image and environmental data, described in the next section.

\section{Agricultural Distributed Decision Framework}
\label{sec:method}

The ADDF is comprised of two distinct functional capabilities: the processing of NDVI image data to compute the evolution of point-in-time crop growth metrics and the ability to learn when to raise a call-to-action, believing that processed image data indicates immediate or impending crop stress. The former capability requires collecting NDVI image data, computing a moving average pixel sufficient metric, and identifying deviating segments of crop fields. The latter capability introduces a reinforcement learning technique for exploring policies that, based on continuous-value observations, trigger a call-to-action.

\subsection{Processing Image Data}
\label{sec:img}

As introduced in Sec.~\ref{sec:bg_img}, composing metrics of growth patterns via NDVI imaging is a popular methodology. Unfortunately, much of the contemporary body of work leverages retrospective curve-fitting as a departure point. As our domain requires that immediate action be taken (i.e. at various points \textit{during} the growing season), we instead opt to develop a sufficient statistic computed from two or more NDVI images taken at potentially diverse intervals by proportionally merging error values.

Two distinct types of image processing using this methodology are required. The first task is simple: two aligned NDVI images, taken from highly similar perspectives (L3 and some L2 tasks), are compared via image difference with $n$-square-pixel approximation. Due to possible slight deviations of the images (due to erroneous alignment or evolution of crop size), the approximation, which takes the average index of an $p\times p$ square, can be used to alleviate information loss.

The second, more complex methodology required is analyzing images of crops from dramatically different angles, comprising some of L2's tasks and nearly all of L1's. In this case, we instead create a generate a distribution of the NDVI image coloration of a crop and compare the distribution to average coloration taken by the layer. In this work, we focus on the first task and the theory of composing a heterogeneous team, leaving this task and online experiments for future work.

\begin{figure} [htb]
\centerline{\includegraphics[width=8cm]{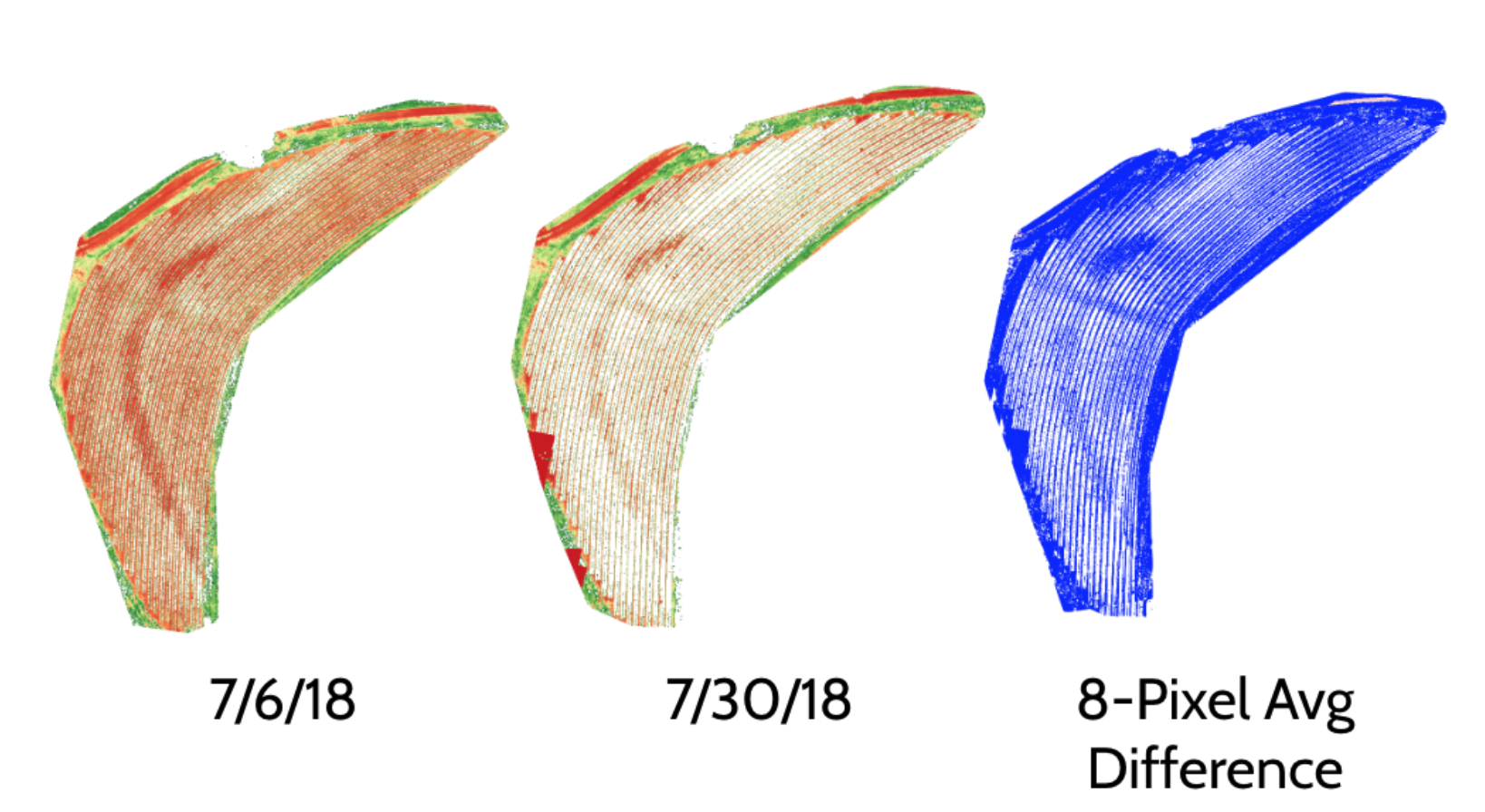}}
\caption{Comparison of two NDVI peanut field images taken over a 3 week period with $8\times 8$ pixel approximation.}
\label{fig:rmsd}
\end{figure}

Figure~\ref{fig:rmsd} shows a trivial implementation of a single pairwise comparison of two images taken 3 weeks apart of our domain's peanut field in the late 2017 growing season. Interestingly, the results are quite telling. Known to the field workers, three standout features are identifiable via this comparison: (1) the field is bisected by two lines as it was originally 4 fields, (2) near the bisection point significant crop erosion is occurring, and (3) the west, and particularly northwest, sector of the field is beset by damage from grazing deer.

\begin{algorithm}[!h]
  \caption{ADDF: NDVI image difference method}
  \label{alg:avg}
  \begin{algorithmic}[1]
    \REQUIRE Two NDVI images, $a, b$; image dimensions $x, y$; pixel approximation parameter $p$
    \STATE $i\gets 0$
    \STATE $j\gets 0$
    \STATE $diff\gets \emptyset$
    \WHILE{$i<x$}
    \WHILE{$j<y$}
    \STATE $a_{i,j}\gets$ average index of pixels $(i,j)$ to $(i+p,j+p)$ for image $a$
    \STATE $b_{i,j}\gets$ average index of pixels $(i,j)$ to $(i+p,j+p)$ for image $b$
    \STATE $diff_{i,j}\gets \min[(a_{i,j}-b_{i,j}),0]$
    \STATE $i\gets i+p$
    \STATE $j\gets j+p$
    \ENDWHILE
    \ENDWHILE
    \RETURN $diff$
  \end{algorithmic}
\end{algorithm}

Algorithm~\ref{alg:avg} annotates the relatively simplistic process of generating a difference sample between two NDVI images. The resultant matrix approximates the index across a $p$ square pixels, the set of which becomes a list of pairwise comparisons. We are interested in $p$-size sectors that, over the growing season, have very low variance, as it indicates that either (a) crops are not growing or are dying quickly, or (b), at the near infrared portion of the spectrum, low reflectance indicates low crop health from water stress.

\begin{algorithm}[!h]
  \caption{ADDF: Variance estimation}
  \label{alg:update}
  \begin{algorithmic}[1]
    \REQUIRE List of pairwise difference matrices $d$; individual diff matrix size $m,n$
    \STATE Variance matrix $v\gets m\times n$ matrix instantiated to $0$
    \IF{$|d|>1$}
    \FOR{$i$ in $[0,m]$; $j$ in $[0,n]$}
    \STATE $v_{i,j}=VAR\{d^k_{i,j}\}_{k=0}^{|d|}$, the variance of the approximate pixel at $i,j$ for all image differences
    \ENDFOR
    \STATE Normalize all values in $v$ between $0$ and $1$
    \ENDIF
    \RETURN $v$
  \end{algorithmic}
\end{algorithm}

As we are primarily interested in disproportionate changes in reflectance and growth, instead of examining pure average indexes across the pairwise differences of images, we focus on the variance of those images. Following the matrix generation in Alg.~\ref{alg:avg}, Alg.~\ref{alg:update} computes the variance across all pairwise differences and then normalizes relative to the highest variance. When analyzing the resultant variance matrix, lower values are more concerning than higher ones.

As a last step, since the resultant matrix serves as rudimentary image data where lower variance is represented as higher intensity pixels. In order to combat variance diffuseness, which is caused by crops separated by rows of soil, we crop the image to the field and apply a Gaussian blur where $\sigma = 2.5$. We then segment the image via straightforward applications of image segmentation algorithms, such as K-means~\cite{kmeans1,kmeans2}, to create afflicted sectors of the crop field. These sectors, based on their average variance, are then decision moments for the agents described in the following sections.

\subsection{Composing a Heterogeneous Team}
\label{sec:team}

As introduced in Sec.~\ref{sec:background}, we have 4 layers of sensor modalities, each with the capability to take either image or environmental data at varying resolutions. We are tasked with representing these layers, then, as a team, with the common goal of identifying a stress while balancing (a) speed of identification and (b) accuracy of the eventual categorization.

\begin{figure} [htb]
\centerline{\includegraphics[width=8cm]{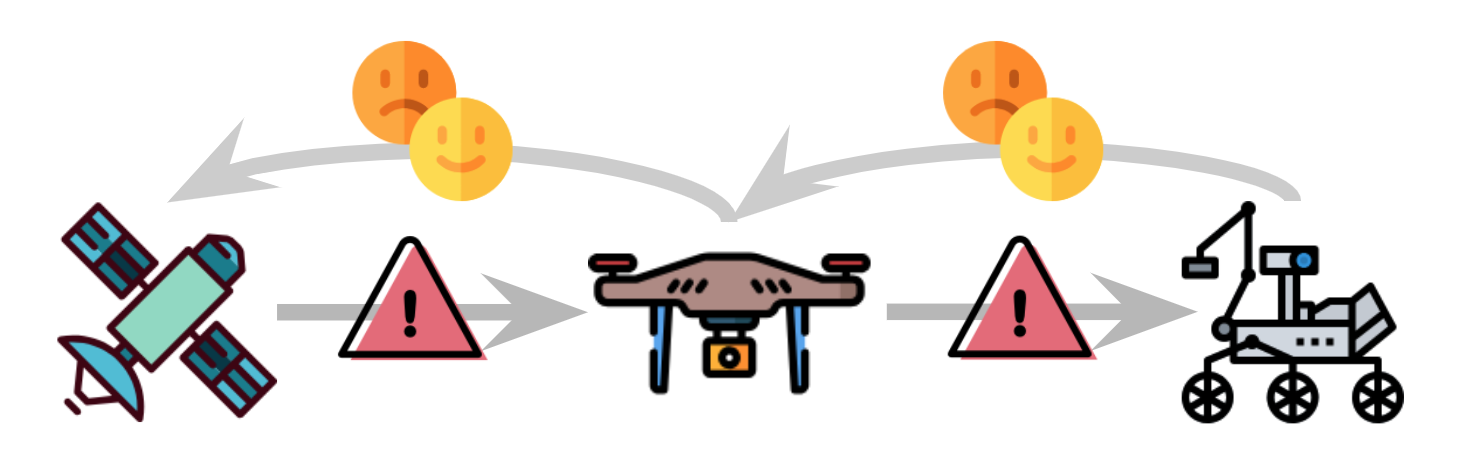}}
\caption{Image demonstrating the call-to-action and reinforcement process of layers 3, 2, and 1.}
\label{fig:rl}
\end{figure}

Casting this problem as decision-theoretic is rather straightforward. Each layer is solving an independent game formulation, though the state of the environment is (largely) identical for each of them. Since the eventual categorization is used to inform the decisions of higher level layers, we adopt the perspective of \textit{reinforcement learning}. Borrowing from Perkins' MCES-P, we explore a set of policies mapping call-to-actions to real-valued observations of the variance across images of a crop field. Figure~\ref{fig:rl} presents a visual representation of the actions of layers and the propagation of feedback.

We first present the general POMDP frame for layers 3, 2, and 1. The problem an individual agent faces is defined as a tuple $ADDF_i^L=\langle S,A,T,\Omega,O,R \rangle$, where $L$ and $i$ refers to the layer and agent respectively. $A^L$ refers to actions taken at this layer, and $A^{L-1}$ refers to the eventual action of the subsequent layer. Level 0 has no agency, represented in our framework as additional state observation information.

\begin{itemize}
\item $S$: the distribution of stress and agent location over a multi-row and column large-scale crop field
\item $A$: the set of actions, uniquely defined per layer
\begin{itemize}
    \item \textbf{L3}: take low-resolution images of the entire field on rare occasions
    \item \textbf{L2}: move; take image of a field or section
    \item \textbf{L1}: move; take image of an individual plants, leaves and fruit
\end{itemize}
\item $T=S\times A\times S$: state transition function dependent on agent movement
\item $\Omega$: the set of observations. Each agent receives information from level 0 and information as to how each sector/crop deviates from expected image data using Alg.~\ref{alg:update}. We discretize to levels of severity via clustering.
\item $O=S\times A$: the observation function, mapping observations to actions dependent on the state
\item $R=S\times A^{L}\times (A^{L-1})\rightarrow \mathcal{R}$: the reward function, dependent on the agents action and the decision made by the subsequent layer. The exception is at layer 1, which makes the final decision on the presence of a stress.
\end{itemize}

\begin{figure} [htb]
\centerline{\includegraphics[width=8cm]{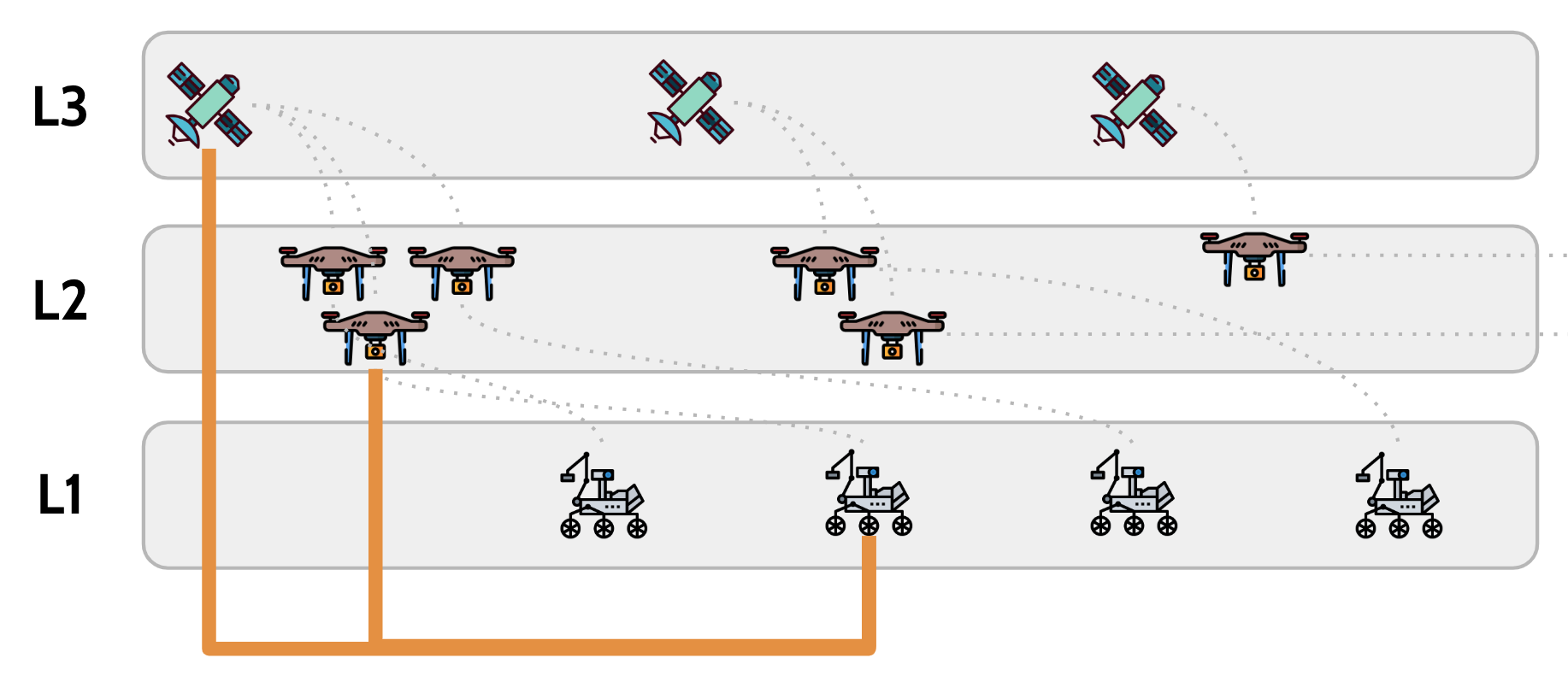}}
\caption{Layers 3 and 2 experience delayed reinforcements when a new game is started before subsequent layers make a decision.}
\label{fig:delayed}
\end{figure}

Examining the reward function $R$ raises an interesting caveat of our domain: reinforcements must be delayed due to relying on subsequent layer categorization, and games are played in parallel even within each layer. Since each game is a single horizon, and the policies that are learned are memory-less and reactive, this only means that games may be resolved outside the order they were played in. Figure~\ref{fig:delayed} demonstrates how a layer 3 agent may begin playing a game before a previous decision was reinforced.

Here we present the algorithms for ADDF. We generally require two flavors of ADDF: a high level agent that may tackle several sectors at once (such as for layer 3 and, sometimes, layer 2) and a lower level, sector- and crop-specific agent (layer 2 and 1). The high level agent creates multiple decision points for the lower agent, which often must tackle the sectors one-by-one.

\begin{algorithm}[!h]
  \caption{ADDF}
  \label{alg:addf}
  \begin{algorithmic}[1]
    \REQUIRE Maximum number of trajectories, $k$, to explore; sectors $S$; number of agents $Z$; observation and action spaces $O,A$ for each agent
    \FOR{$i\in (2,Z)$} \label{line:init}
    \STATE $Q^Li_{o,a}\gets 0$ for all $(O_i,A_i)$
    \STATE $c^Li_{o,a}\gets 0$ for all $(O_i,A_i)$
    \STATE $\pi^Li\gets $random initial policy
    \ENDFOR
    \REPEAT
    \STATE Generate call to actions $\vec{\tau}^{L3}=L3(\pi^{L3},k,S,c^{L3})$ \label{line:L3}
    \STATE Generate call to actions and collect rejections $\langle \vec{\tau}^{L2}, \vec{r}^{L2} \rangle =L2(\pi^{L2},k,S,c^{L2},\vec{\tau}^{L3})$ \label{line:L2}
    \STATE Collect classifications $\vec{r}^{L1} =L1(\vec{\tau}^{L2})$
    \STATE $\vec{r} = \vec{r}^{L1} + \vec{r}^{L2}$
    \STATE $Update(\vec{Q},\vec{c},\vec{\pi},\vec{r},\vec{\tau},Z)$ \label{line:learn}
    \UNTIL{End of growing season}
  \end{algorithmic}
\end{algorithm}

We begin by covering the high-level rotation of actions as a crop season progresses, defined in Alg.~\ref{alg:addf}. Line~\ref{line:init} initializes the Q-table, counts, and policies for each layer. Line~\ref{line:L3} runs the call-to-action generation for Layer 3, leveraging its current learned policy. Layer 2 is a bit different. While line~\ref{line:L2} also generates calls to action, any sector it considers without stress is served as stimuli for Layer 3. Layer 1 only generates stimuli. Line~\ref{line:learn} then updates all Q-values and transforms layer policies that have a new best action using reward stimuli.

As an important point, we omit the constraint of parallelism in the description of Alg.~\ref{alg:addf} for the purpose of brevity and clarity. Line~\ref{line:L2}, for example, doesn't occur every iteration. This is accomplished instead by creating a queue from line~\ref{line:L3} and iteratively executing line~\ref{line:L2} until the queue is empty, illustrated in Fig.~\ref{fig:delayed}. Line~\ref{line:learn} only executes when Layer 1 or 2 completes a task.

\begin{algorithm}[htbp]
  \caption{$L3(\pi,k,S,c)$}
  \label{alg:l3}
  \begin{algorithmic}[1]
    \REQUIRE Policy $\pi$; a maximum number of trajectories, $k$, to explore; vector of potentially stressed sectors $S$; counts $c$
    \STATE Choose random action $a$ and observation $o$
    \STATE Generate $\pi'$ as $\pi \gets (o, a)$
    \STATE Create variance samples $X$ from $S$ following Alg.~\ref{alg:update}
    \STATE Generate trajectories $\hat{\tau} = \{\tau_x\}_{x=0}^{|X|}$, according to $\pi'$
    \STATE Update counts $c_{o,a}\gets c_{o,a} + 1$
    \STATE Extract call-to-actions for subsequent layers, $\vec{\tau}=\{\tau \in \bar{\tau} : \tau_a > 0\}$ \label{line:extract}
    \RETURN $\vec{\tau}$
  \end{algorithmic}
\end{algorithm}

Algorithm~\ref{alg:l3} defines the L3 policy exploration and execution process. As in Alg.~\ref{alg:MCESP}, we select a random observation-action pair to explore, creating transformed policy $\pi'$. By taking variance samples of target sectors, L3 generates observations, which it then acts on using the transformed policies. For those actions that indicate a stress, L3 returns a call-to-action.

Since information must be passed between layers, we define $\tau$ differently than canonical MCESP. Here, each element in $\tau^{Li}$ includes the observation, action, and sector index ($o,a,s$) of that layer, omitting rewards. When we return rewards back to preceding layers via $\vec{r}$, the preceding layer then knows which observation and action to update.

The algorithm for Layer 2 is omitted, as it differs from Layer 3 only in that it returns trajectories from $\vec{\tau}^{L3}$ as a negative reward if it fails to detect a stress in that sector, similar to Layer 1's line~\ref{line:reject} below, albeit only negative rewards.

\begin{algorithm}[!h]
  \caption{$L1(I)$}
  \label{alg:l1}
  \begin{algorithmic}[1]
    \REQUIRE Call-to-actions $I$
    \STATE $\vec{r}\gets\emptyset$
    \FOR{States from call-to-actions $s\in I$}
    \STATE Create variance samples $x$ from $i(s)$ following Alg.~\ref{alg:update}
    \STATE Classify non-uniformity $r=Classify(x)$ \label{line:classify}
    \STATE Dispense rewards $\vec{r} = APPEND(\vec{r},(s,r))$ \label{line:reject}
    \ENDFOR
    \RETURN $\vec{r}$
  \end{algorithmic}
\end{algorithm}

In our formulation, Layer 1 is considered objective in its classification, and generates stimuli for the other two (or more) layers. Like the previous layers, it generates an image of a particular crop, but if it detects significant non-uniformity, it immediately classifies the sector as stressed. 

\begin{algorithm}[!h]
  \caption{$Update(\vec{Q},\vec{c},\vec{\pi},\vec{r},\vec{\tau},Z,k)$}
  \label{alg:rlupdate}
  \begin{algorithmic}[1]
    \REQUIRE Q-tables $\vec{Q}$; count vectors $\vec{c}$; current policies $\vec{\pi}$; rewards $\vec{r}$; call-to-actions $\vec{\tau}$; number of agents $Z$; trajectory requirement count $k$
    \FOR{Agent $i\in (2,Z)$}
    \FOR{Agent's call-to-actions $t\in \vec{\tau}^{Li}$}
    \STATE Get reward for call-to-action $r = \vec{r}(t(s))$ \label{line:feedback}
    \STATE $Q^{Li}_{o, a} \gets (1-\alpha(c^{Li}_{o, a})) \cdot Q^{Li}_{o, a} +\alpha(c^{Li}_{o,a}) \cdot r$
    \IF {$\max_{a'} Q^{Li}_{o, a'} > Q^{Li}_{o, \pi^{Li}(o)} + \epsilon(k,c^{Li}_{o,a}, c^{Li}_{o,\pi^{Li}(o)})$}
    \STATE $\pi^{Li}(o) \gets a'$
    \STATE $c^{Li}_{o, a} \gets 0$ for all ${o,a}$
    \ENDIF
    \ENDFOR
    \ENDFOR
  \end{algorithmic}
\end{algorithm}

Each iteration is concluded with an update of the Q-table. As in MCESP, if the sample complexity $k$ is satisfied, then transformations are possible, if the agent has learned enough about each potential policy.

Our domain differs from predominant exploration of POMDP domains in that the observation function is continuous. That is, we receive real-valued differentials between expected crop growth when processing image data. However, via utilizing lossless conversion to observation space partitions~\cite{hoey}, we can solve this continuous problem discretely, even considering layer 0 continuous-value environmental metadata. 

In layers that contain multiple agents (excluding layer 3 and 0), since the problem requires potentially reacting to multiple call to actions, we will utilize a generalization of the POMDP that solves for instant-communication team play, the Multiagent POMDP~\cite{boutilier}. The MPOMDP is an interesting formulation, representing the dynamism of the decision problem by individually capturing the characteristics of each component of the system (in this case, the layers). However, it is well-known that an MPOMDP can be directly converted to an equally expressive POMDP and solved as such~\cite{amato}. The value is in its interpretability, which doesn’t affect the complexity to solve it.

We represent a layer’s homogeneous team as a tuple $ADDF^L=<Z,S,\vec{A},T,\Omega,O,R>$, with new or modified frame elements:
\begin{itemize}
\item $Z$: the set of agents
\item $\vec{A}={A^L_z}_{z=1}^Z$: the set of all agents’ actions
\item $O=S\times \vec{A}$: the team observation function, where the state and the team’s actions result in a single, shared observation
\end{itemize}

Utilizing our reinforcement learning perspective, we then learn the optimal policy for each layer as a Monte Carlo solution to the (M)POMDPs~\cite{thrun,perkins}.

\subsection{Tackling the Hyper-Conservative Local Optima}
\label{sec:heuristics}

One complication of our methodology is that layers are incentivized to be highly conservative in their call-to-actions. Since the presence of a stress is a rare event \textit{vis a vis} normal growth patterns, learning is inherently biased to rejecting the stress~\cite{rare_event}. Besides dramatically overweighting rewards that indicate stress, we propose two options: forced exploration even under rejection and random exploration in layer 1.

The first option considers reserving a few call-to-actions at each interval (demonstrated in Fig.~\ref{fig:delayed}) to create an objective for subsequent layers to explore near-accepted call-to-actions. For example, even if layer 3 rejects the notion of a stress in a particular sector, it will still inform layer 2 of a stress in that sector (along with other positively-identified stresses), a strategy similar to random policy exploration~\cite{wu13}. This may be added to Alg.~\ref{alg:l3} by adding rejected sectors to the call-to-actions $\tau$ with exponentially-decaying weight $w(\tau,m,i)=\frac{m}{m+|\tau|+i}$ for each $i$th rejected sector where $m$ controls the steepness of the exponential decay. This methodology is explored in Sec.~\ref{sec:results}.

The second option capitalizes on potential dead-time in the lower layers. When not exploring the crop field, we can opt to set layer 1 and 2 to an exploration mode, where they peruse the field randomly. When a stress is identified, a call-to-action may be simulated sequentially through the entire system and immediately rewarded appropriately, creating a positive sample to learn from. We reserve exploration of this methodology for real-world experiments in future work.

\section{Experimental Results}
\label{sec:results}

We propose two toy implementations of our problem domains to test the approaches in Secs.~\ref{sec:img} and~\ref{sec:team}. We first examine the performance of our image segmentation technique for identifying stressed crop areas by utilizing actual AUAV images collected across several weeks of a growing season. We then introduce a two-layer POMDP implementation of our ADDF framework with a simulated toy environment, showing the effectiveness of learning a reward function \textit{vis a vis} a subsequent layer's decisions. In practice, Sec.~\ref{sec:results_image} produces inputs for Sec.~\ref{sec:results_team}, but we perform separate experiments in this work.

\subsection{Identifying Phenotypic Stress}
\label{sec:results_image}

Through technology supplied to the Department of Entomology at the University of Georgia, Tifton campus, we collected 5 images, each separated by 7 to 9 days, in July and August of 2017 of a peanut field. These images were collected by a 3DR Solo aerial drone indexed using NDVI at a resolution of 1-3 centimeters per pixel. The resultant files were encoded as Tagged Image File Format (TIFF), sized around 20 megabytes each.

Since these images took place in a previous growing season, the department is aware of several stresses that occurred during the growing season. First, the peanut field used to be four separate fields, bisected by two roads, and a sector just north of the east-west road suffered from stress due to soil erosion.

We experimented with several parameter settings for $p$, $\sigma$, and, in the case of K-means, $k$. Though we hand-selected parameters for the final result in this section, we hypothesize that performing gradient exploration methodologies, parameters could be fit by optimizing for fit in the final segmentation technique (such as the elbow method for K-means~\cite{elbow}.

\begin{figure}[!htb]
    \centering
    \begin{minipage}{.25\textwidth}
        \centering
        \includegraphics[width=0.9\textwidth]{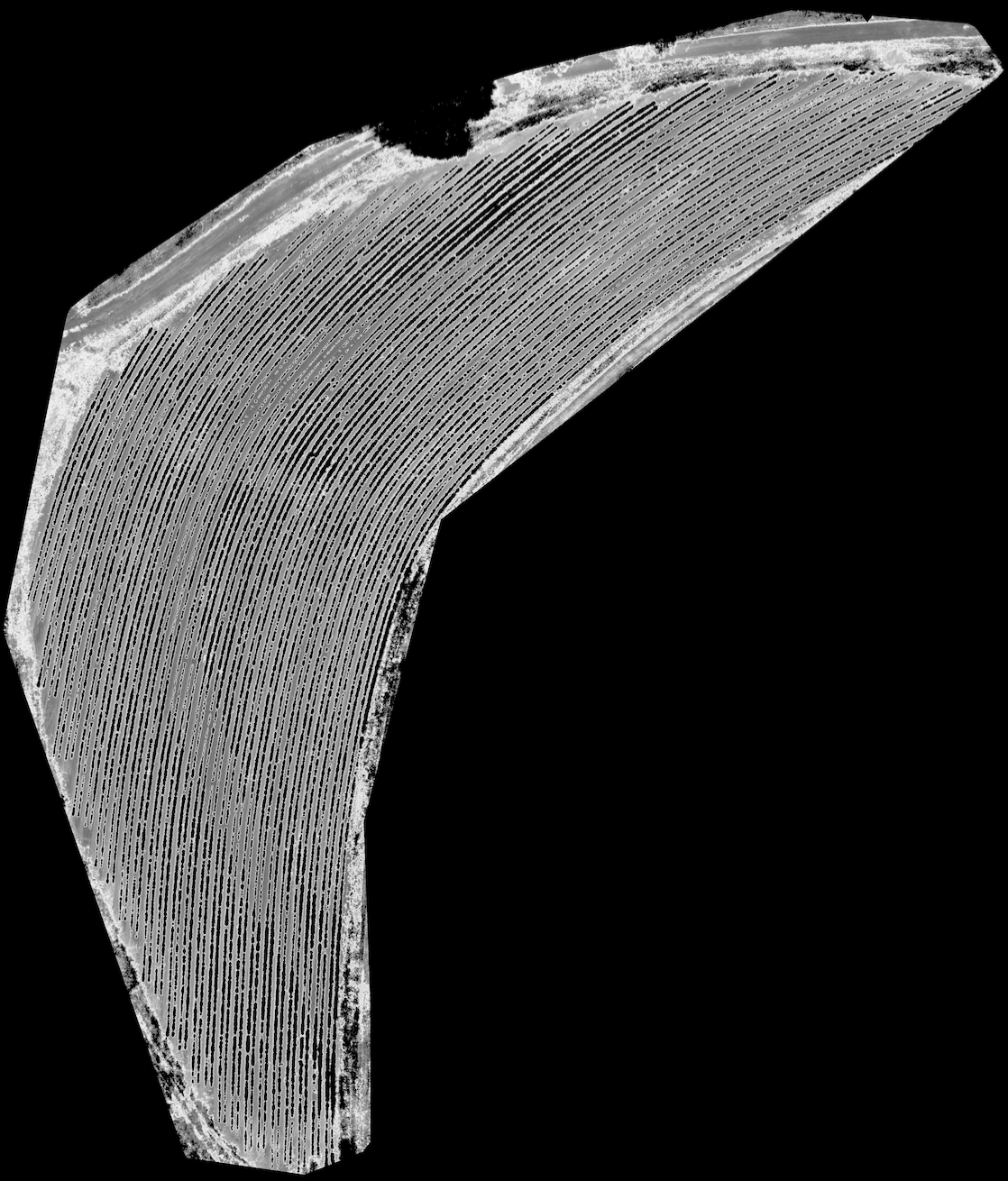}\\
        {$(a)$}
    \end{minipage}%
    \begin{minipage}{0.25\textwidth}
        \centering
        \includegraphics[width=0.9\textwidth]{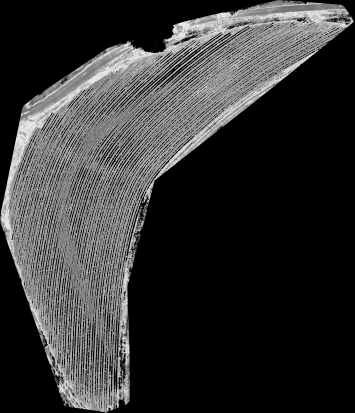}\\
        {$(b)$}
    \end{minipage}
    \caption{Illustration of applying Alg.~\ref{alg:avg} with $p>1$ to an original image $(a)$ resulting in an approximate image $(b)$.}
    \label{fig:approx}
\end{figure}

For Alg.~\ref{alg:avg}, we set $p=12$, converting original image dimensions of around $4900 x 4200$ to $620 x 530$. Figure~\ref{fig:approx} shows the effect of approximation, resulting in significantly less diffuseness between crop rows, though some is retained. Since the result of Gaussian blurring can lose too much information to compare between images, we wait until the final comparative image before applying it. Next, we compared the differences between two subsequent crop field images.

\begin{figure}[!htb]
    \centering
    \begin{minipage}{0.25\textwidth}
        \centering
        \includegraphics[width=0.7\textwidth]{Figs/approx1}\\
        {$(a)$}
    \end{minipage}%
    \begin{minipage}{0.25\textwidth}
        \centering
        \includegraphics[width=0.7\textwidth]{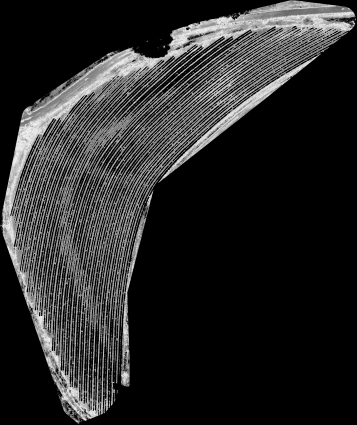}\\
        {$(b)$}
    \end{minipage}
    \begin{minipage}{0.01\textwidth}
        $\Large{\rightarrow}$
    \end{minipage}
    \begin{minipage}{0.25\textwidth}
        \centering
        \includegraphics[width=0.7\textwidth]{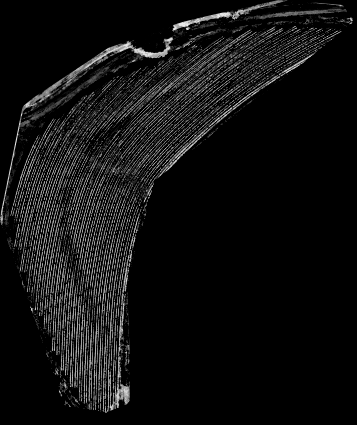}\\
        {$(c)$}
    \end{minipage}
    \caption{Result after full completion of Alg.~\ref{alg:avg}, where the difference of two subsequent images ($(a)$ July 6th and $(b)$ July 13th) is computed, resulting in $(c)$.}
    \label{fig:diff}
\end{figure}

In Fig.~\ref{fig:diff} we see the result of the first $diff$ performed in our sample set for the first two dates. It confirms what we know about this data set: the north side of the field (which is darker, indicating less change) is performing significantly worse than the rest of the field, and roads bisecting the field are clearly visible.

\begin{figure}[!htb]
    \centering
    \begin{minipage}{0.25\textwidth}
        \centering
        \includegraphics[width=0.7\textwidth]{Figs/diff1}\\
        {$(a)$}
    \end{minipage}%
    \begin{minipage}{0.25\textwidth}
        \centering
        \includegraphics[width=0.7\textwidth]{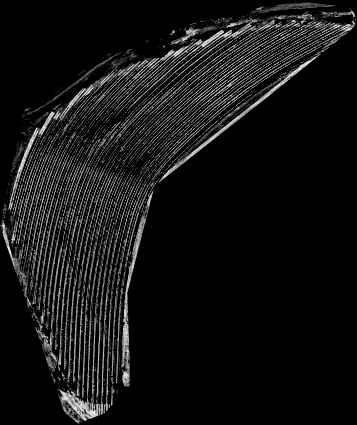}\\
        {$(b)$}
    \end{minipage}
    \begin{minipage}{0.01\textwidth}
        $\Large{\rightarrow}$
    \end{minipage}
    \begin{minipage}{0.25\textwidth}
        \centering
        \includegraphics[width=0.7\textwidth]{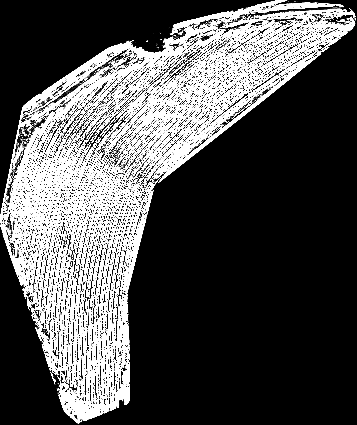}\\
        {$(c)$}
    \end{minipage}
    \caption{Calculating the variance between differences computed for $(a)$ July 6th and 13th and $(b)$ July 13th and 21st, resulting in variance map $(c)$.}
    \label{fig:var}
\end{figure}

\begin{figure}[!htb]
    \centering
    \begin{minipage}{0.25\textwidth}
        \centering
        \includegraphics[width=0.7\textwidth]{Figs/var}\\
        {$(a)$}
    \end{minipage}%
    \begin{minipage}{0.01\textwidth}
        $\Large{\rightarrow}$
    \end{minipage}
    \begin{minipage}{0.25\textwidth}
        \centering
        \includegraphics[width=0.7\textwidth]{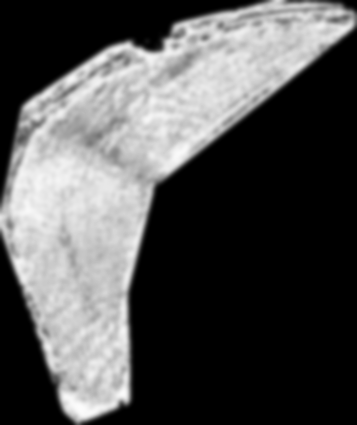}\\
        {$(b)$}
    \end{minipage}
    \begin{minipage}{0.01\textwidth}
        $\Large{\rightarrow}$
    \end{minipage}
    \begin{minipage}{0.25\textwidth}
        \centering
        \includegraphics[width=0.7\textwidth]{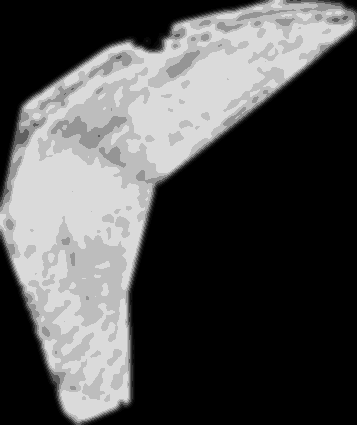}\\
        {$(c)$}
    \end{minipage}
    \caption{To identify affected sectors, we take an $(a)$ variance image, apply $(b)$ Gaussian blur, and then $(c)$ K-means.}
    \label{fig:kmeans}
\end{figure}

Examining the images in Fig.~\ref{fig:var} corroborates the growing concern from $diff$s computed on July 13th and 21st that the middle mass of the crop field is not producing. One drawback of this methodology is that the stress is certainly visible by the $diff$ on July 13th (recovering slightly by the 21st). After applying Alg.~\ref{alg:update}, since the variance image is still too diffuse for effective image segmentation, we apply a Gaussian blur and then segment, demonstrated in Fig.~\ref{fig:kmeans}.

Completing the approximation and segmenting technique then results in highly defined and contoured areas, show in Fig.~\ref{fig:kmeans}$.c$ for July 21st. 3 distinct layers are clearly visible (though the number of actual layers are determined by the input $k$ for the K-means method, $10$ in our experiments). The center road and previously mentioned erosion due north of the center of mass are both illustrated as very severe, with less severe sectors in the south and northeast.

\begin{figure}[!htb]
    \centering
    \begin{minipage}{0.25\textwidth}
        \centering
        \includegraphics[width=0.7\textwidth]{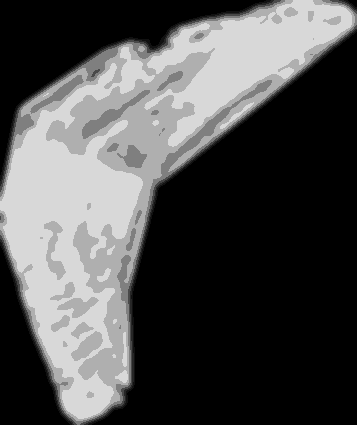}\\
        {$(a)$}
    \end{minipage}
    \begin{minipage}{0.25\textwidth}
        \centering
        \includegraphics[width=0.7\textwidth]{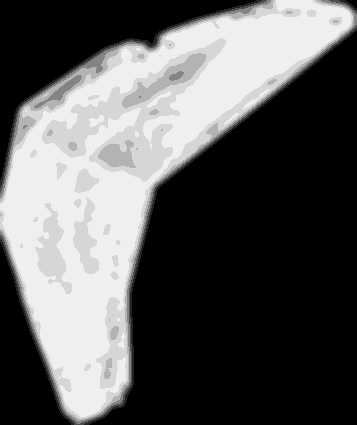}\\
        {$(b)$}
    \end{minipage}
    \caption{K-means processing for $(a)$ July 30th and $(b)$ August 11th}
    \label{fig:final_images}
\end{figure}

We complete this section by illustrating the K-means processing for the final two dates in our test set. Over the course of the first month and a half, we see that growth is stymied after our first completion and the problem worsens in the north side of the field, though it improves in the south. By the last date, August 11th, the field is significantly healthier, though the most severe problem areas from July 13th persist.

\subsection{Solving Inter-Team Objectives}
\label{sec:results_team}

\begin{figure} [htb]
\centerline{\includegraphics[width=8cm]{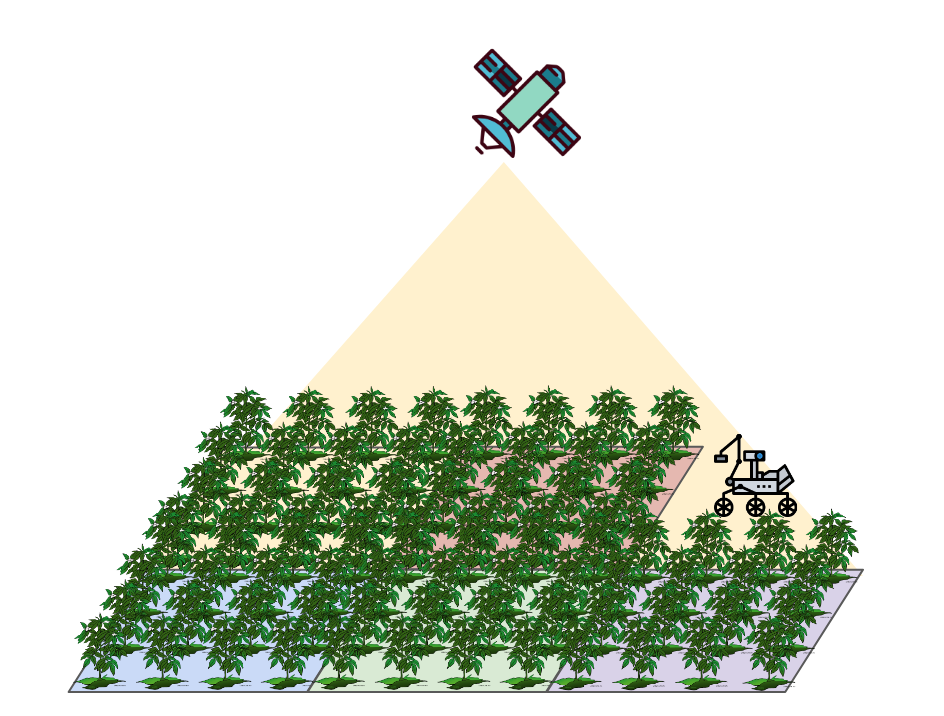}}
\caption{The toy crop field problem, where a fast and slow agent team up to learn crop stress probabilities.}
\label{fig:problem_domain}
\end{figure}

We test the effectiveness of the ADDF algorithm in a toy implementation of the crop field problem~\footnote{Code available on github at https://github.com/quinngroup/addf$\_$pomdp}. In our experiment, two agents form a two-layer team, where the first agent (referred to as the "fast" agent) takes a low-precision image of an entire crop field, which contains 5 sectors, once every 3 days. The second, "slow" agent can collect image data only for a single sector, but can do it once a day. To simulate a L1 classifier, we use an oracle that returns the true state of the sector after each agent acts on a sector. Each agent receives one of $|O|$ observations, each correlating to a confidence of the stress, from low to high. We illustrate the domain in Fig.~\ref{fig:problem_domain}, where the fast and slow agents are represented by a satellite and AUGV, respectively.

The simulator generates a stress in each sector at the beginning of a growing season, which has $89$ days, with a $50\%$ probability. Early in the season, the stress of the state is unstable, having a $50\%$ chance to change status. This likelihood decreases exponentially each subsequent day. The fast agent gets the maximally correlated observation of the true stress of the environment (either $o=0$ or $o=|O|$ for the lack or presence of a stress, respectively) with $70\%$ probability, while the slow agent receives it with $85\%$ probability. Incorrect observations are received with the remaining probability with exponential weight towards the correct classification. For example, the slow agent's observations are received with probability $\{o0=0.8, o1=0.1, o2=0.05\}$ when no stress is present.

We test ADDF with configurations varying the number of observations and the inclusion of the heuristic in Sec.~\ref{sec:heuristics}, which adds rejected sectors to call-to-actions with exponentially decaying probabilities. Each configuration is tested with 500 seasons. We begin by setting the baseline with a canonical Q-learning technique, which does not perform policy exploration and always exploits the current highest action-value for an observation.

\begin{table}[h]
	\centering
	\scalebox{1}{%
		\begin{tabular}{|c|c|c|c|c|c|c|c|}
			\hline
			\multirow{3}{*}{\textbf{$|O|$}} & \multirow{3}{*}{\textbf{Agent}} & \multicolumn{5}{c|}{\textbf{Accuracy}} \\
			\cline{3-7}
			& & \multicolumn{2}{c|}{\textbf{True}} & \multicolumn{2}{c|}{\textbf{False}}  & \multirow{2}{*}{\textbf{Overall}}\\
			\cline{3-6}
			& & \textbf{Positive} & \textbf{Negative} &  \textbf{Positive} & \textbf{Negative} &\\
			\hline \hline
			\multirow{2}{*}{\textbf{3}} & \textbf{Fast} & 5 & 37,416 & 2 & 37,577 & $49.9\%$\\
			\cline{2-7}
			& \textbf{Slow} & 2 & 3 & 1 & 1 & \st{$71.4\%$} \\
			\hline
			\multirow{2}{*}{\textbf{5}} & \textbf{Fast} & 64 & 37,695 & 3 & 37,238 & $50.3\%$\\
			\cline{2-7}
			& \textbf{Slow} & 35 & 4 & 1 & 27 & \st{$56.3\%$} \\
			\hline
		\end{tabular}
	}
	\caption{Q-learning baseline, varying the observation and sector counts.}
	\label{tbl:qbaseline}
	\vspace{-0.9em}
\end{table}

As noted in Sec.~\ref{sec:heuristics}, the relative rarity of positive stimuli for negative events causes Q-learning to quickly converge to always rejecting the presence of a stress in a sector. Therefore, Q-learning performs essentially the same as the random baseline. With almost no samples to learn from, we strike the essentially random accuracy of the slow agent. We then test ADDF in this domain with trajectory limit $k=500$.

\begin{table}[h]
	\centering
	\scalebox{1}{%
		\begin{tabular}{|c|c|c|c|c|c|c|c|}
			\hline
			\multirow{3}{*}{\textbf{$|O|$}} & \multirow{3}{*}{\textbf{Agent}} & \multicolumn{5}{c|}{\textbf{Accuracy}} \\
			\cline{3-7}
			& & \multicolumn{2}{c|}{\textbf{True}} & \multicolumn{2}{c|}{\textbf{False}}  & \multirow{2}{*}{\textbf{Overall}}\\
			\cline{3-6}
			& & \textbf{Positive} & \textbf{Negative} &  \textbf{Positive} & \textbf{Negative} &\\
			\hline \hline
			\multirow{2}{*}{\textbf{3}} & \textbf{Fast} & 26,068 & 34,032 & 3,671 & 11,229 & $80.1\%$\\
			\cline{2-7}
			& \textbf{Slow} & 13,067 & 11,895 & 4,068 & 709 & $83.9\%$ \\
			\hline
			\multirow{2}{*}{\textbf{5}} & \textbf{Fast} & 20,756 & 36,737 & 3,732 & 14,775 & $75.6\%$\\
			\cline{2-7}
			& \textbf{Slow} & 12,052 & 7,612 & 701 & 4,123 & $80.3\%$ \\
			\hline
		\end{tabular}
	}
	\caption{ADDF with $k=500$.}
	\label{tbl:mcesp}
	\vspace{-0.9em}
\end{table}

ADDF performs dramatically better than the Q-learning baseline, achieving over $80\%$ accuracy for the slow agent and 3-obervation fast agent, performing relatively worse with more observations for the fast agent. The fast agent performs comparatively worse than the slow agent likely due to the increased noise in its observation function. These numbers are quite close to the theoretical maximum, considering noise for the environment.

Both the baseline and ADDF can potentially benefit from increased workload. The baseline rejects nearly every sector, and, referring to Tbl.~\ref{tbl:mcesp}, it is clear the slow agent is not given enough decision points. For example, when $|O|=3$, the fast agent generates just under 60 calls to the slow agent per season, when it can work 90. Without lowest-level agents working every day, many stressed crops will not be identified. We show the results for both the baseline and MCESP with the workload heuristic in Sec.~\ref{sec:heuristics} with $m=5$.

\begin{table}[h]
	\centering
	\scalebox{1}{%
		\begin{tabular}{|c|c|c|c|c|c|c|c|}
			\hline
			\multirow{3}{*}{\textbf{Method}} & \multirow{3}{*}{\textbf{$|O|$}} & \multirow{3}{*}{\textbf{Agent}} & \multicolumn{5}{c|}{\textbf{Accuracy}} \\
			\cline{4-8}
			& & & \multicolumn{2}{c|}{\textbf{True}} & \multicolumn{2}{c|}{\textbf{False}}  & \multirow{2}{*}{\textbf{Overall}}\\
			\cline{4-7}
			& & & \textbf{Positive} & \textbf{Negative} &  \textbf{Positive} & \textbf{Negative} &\\
			
			\hline \hline
			
			\multirow{4}{*}{\rotatebox{90}{\textbf{Baseline}}} & \multirow{2}{*}{\textbf{3}} & \textbf{Fast} & 30552 & 26445 & 14202 & 3801 & $76\%$ \\
			\cline{3-8}
			& & \textbf{Slow} & 22238 & 13629 & 5110 & 5023 & $78\%$ \\
			\cline{2-8}
			& \multirow{2}{*}{\textbf{5}} & \textbf{Fast} & 25249 & 28579 & 12067 & 9105 & $71.8\%$ \\
			\cline{3-8}
			& & \textbf{Slow} & 17061 & 15549 & 4927 & 7463 & $72.5\%$ \\
			\hline \hline
			\multirow{4}{*}{\rotatebox{90}{\textbf{ADDF}}} & \multirow{2}{*}{\textbf{3}} & \textbf{Fast} & 30703 & 31355 & 2327 & 10615 & $82.7\%$ \\
			\cline{3-8}
			& & \textbf{Slow} & 24260 & 13488 & 4936 & 1415 & $83.9\%$ \\
			\cline{2-8}
			& \multirow{2}{*}{\textbf{5}} & \textbf{Fast} & 28025 & 31995 & 6239 & 8741 & $80\%$ \\
			\cline{3-8}
			& & \textbf{Slow} & 17865 & 18242 & 3991 & 4898 & $80.2\%$ \\
			\hline
		\end{tabular}
	}
	\caption{The baseline and ADDF algorithms utilizing increased workload heuristics to employ the slow agent more frequently.}
	\label{tbl:heuristic}
	\vspace{-0.9em}
\end{table}

Table~\ref{tbl:heuristic} shows a dramatic performance boost for the baseline but, much more importantly, additionally demonstrates the slow agent working nearly every single available day. While ADDF still outperforms the baseline, it does not show remarkable improvement over the non-heuristic version. However, this isn't the major contribution of the heuristic. The impact is that, since the slow agent works more days, it identifies up to $83\%$ more stressed crops.

\section{Future Work}
\label{sec:future}

Several avenues exist for expanding the prototypical implementation we presented in this work. While we simulate layer 0 information in Sec.~\ref{sec:results_team}, we do not explore the ramifications it may have on image processing in Sec.~\ref{sec:results_image}. Additionally, layer 1 image processing is a completely different, and far more complex, task than the algorithms presented in Sec.~\ref{sec:img}. However, even in real-world experiments, the tasks accomplished by layer 1 may be readily replicated by a plant pathologist.

Image processing may additionally be improved, beyond layer 1, via aligning images via homography. In some recent work, UAV image data may employ homography computation to correct for small alterations, relieving some of the pressure on our approximation technique in Sec.~\ref{sec:img}~\cite{homography}.

Further expanding on future work in Sec.~\ref{sec:img}, there is much possibility for more profound image segmentation techniques that can immediately be leveraged by the ADDF framework. Primarily, performing image segmentation to identify and align subsections across the temporal evolution of the system, performing RMSD on subimages, is a potentially fruitful path, explored recently via deep neural networks~\cite{plant_dnn} as a learned classifier. This would be particularly valuable as a replacement for the proposed layer 1 methodology.

\bibliographystyle{abbrv}
\bibliography{addf}
\end{document}